\documentclass[runningheads]{llncs}

\usepackage{eccv}

\usepackage{eccvabbrv}

\usepackage{graphicx}
\usepackage{booktabs}
\usepackage{neuralnetwork}
\usepackage{tikz}

\usepackage[accsupp]{axessibility}  

\usepackage{hyperref}

\usepackage{orcidlink}

\begin{document}

\title{Enhancing Facial Expression Recognition through Dual-Direction Attention Mixed Feature Networks: Application to 7th ABAW Challenge}

\author{Josep Cabacas-Maso\inst{1}\and
Elena Ortega-Beltrán \inst{1}\and
Ismael Benito-Altamirano\inst{1,2} \and Carles Ventura \inst{1}}


\institute{eHealth Center, Faculty of Computer Science, Multimedia and Telecommunicactions, Universitat Oberta de Catalunya, 08016 Barcelona, Spain
\email{ibenitoal@uoc.edu}, \email{cventuraroy@uoc.edu}\\ \and
MIND/IN2UB, Department of Electronic and Biomedical Engineeering, Universitat de Barcelona, 08028 Barcelona, Spain \\
}

\authorrunning{Josep Cabacas-Maso et al.}
\titlerunning{Enhancing Facial Expression Recognition through DDAMFN}

\maketitle

\begin{abstract}
We present our contribution to the 7th ABAW challenge at ECCV 2024, by utilizing a Dual-Direction Attention Mixed Feature Network (DDAMFN) for multitask facial expression recognition, we achieve results far beyond the proposed baseline for the Multi-Task ABAW challenge. Our proposal uses the well-known DDAMFN architecture as base to effectively predict valence-arousal, emotion recognition, and facial action units. We demonstrate the architecture ability to handle these tasks simultaneously, providing insights into its architecture and the rationale behind its design. Additionally, we compare our results for a multitask solution with independent single-task performance.
\end{abstract}

\section{Introduction}
\label{sec:intro}

Facial emotion recognition has emerged as a pivotal area of research within affective computing, driven by its potential applications in fields ranging from human-computer interaction to psychological research and clinical diagnostics. Since Ekman's classification of human expression faces into emotions~\cite{ekman1978facial}, a lot of studies have emerged, in recent years. Calvo et al.~\cite{calvo2016faces}, Baltrusaities et al.~\cite{baltrusaitis2018openface}, or Kaya et al.~\cite{kaya2020exploring} laid foundational frameworks for understanding facial expressions as a window into emotional states. In contemporary research, Liu et al.~\cite{liu2021deep} and Kim et al.~\cite{kim2023survey} continued to refine and expand these methodologies, by synthesizing insights from cognitive psychology, computer vision, and machine learning, researchers have made significant strides in enhancing the accuracy and applicability of facial emotion recognition systems. Moreover, the integration of valence and arousal dimensions~\cite{soleymani2017deep,zafeiriou2017aff} added depth to the interpretation of emotional states, enabling more nuanced insights into human affective experiences.

Action unit detection~\cite{ekman2003darwin, lucey2010extended} complemented these efforts by parsing facial expressions into discrete muscle movements, facilitating a finer-grained analysis of emotional expressions across cultures and contexts. Such advancements not only improved the reliability of automated emotion recognition systems but also opened the possibility to personalize affective computing applications in fields such as mental health monitoring~\cite{picard2020toward} or user experience design~\cite{zeng2019survey}.

To tackle all these challenges, researchers have explored innovative architectures such as the DDAMFN (Dual-Direction Attention Mixed Feature Network)~\cite{zhang2023ddamfn}. This novel approach integrates attention mechanisms~\cite{hou2021coordinate} and mixed feature extraction~\cite{chen2018mobilefacenets}, enhancing the network's ability to capture intricate details within facial expressions. This architecture shows promising results in multitask challenges, together with other pretrained networks~\cite{savchenko2024hsemotionteam6thabaw}.

There exists a modern day need to create machines capable to comprehend and appropriately respond to human feelings day-to-day on-the-wild applications. This challenge was presented series of comptetitions entitled ``Affective Behavior Analysis in-the-wild (ABAW)'' challenges~\cite{kollias2023abaw2,kollias2023multi,kollias2023abaw,kollias2022abaw,kollias2021analysing,kollias2021affect,kollias2021distribution,kollias2020analysing,kollias2019expression,kollias2019deep,kollias2019face}. For the 7th ABAW challenge, at ECCV 2024, two competitions where presented: first, a competition focused on solving a multi-task classification, focused on valance-arousal, emotion recognition, and action units, and second, a competition focused on compound expression recognition. In this work, we present our approach to the first competition, where we implemented our multi-task version of the DDAMFN architecture.

The competition presented a smaller dataset (s-AffWild2)~\cite{kollias20247th} than in previous challenges (Aff-Wild2)~\cite{kollias20246th}. We proposed a fine-tuning training on the s-AffWild2 dataset to increase the performance of the model for a multitask challenge; plus, we evaluated its performance individually on the different task of the challenge: valence-arousal prediction, emotion recognition, and action unit detection.

\section{Methodology}

\subsection{Dataset curation}

The s-Aff-Wild2 was introduced by the organizers of 7th ABAW challenge~\cite{kollias20247th} as a subset of Aff-Wild2, representing a subset of image frames of the original videos, without any audio data. Only the frames and the annotations for 221,928 images were presented. The dataset presented a preestablished train-validation-test partition: 142,382 instances were introduced for training; 26,876, for validation; and 52,670 where released later on in the challenge as test set. Aproximating, relative partions of: 65\% for training, 12\% for validation and 23\% for test. The frames came with annotations in terms of valence-arousal, eight basic expressions (happiness, sadness, anger, fear, surprise, and disgust), the neutral state, and an `other' category that included affective states not covered by the aforementioned categories. Additionally, the dataset contained annotations for 12 action units: AU1, AU2, AU4, AU6, AU7, AU10, AU12, AU15, AU23, AU24, AU25, and AU26.Així com la comparació dels histogrames de valance-arousal.

Following the instructions provided by the competition organizers we filtered out image frames that contained invalid annotations. We decided to apply an strict criteria to filter out the frames, removing any frame that contained any annotation value outside the specified acceptable range. Specifically, annotation values of -5 for valence/arousal, -1 for expressions, and -1 for action units (AUs) were excluded from consideration in the analysis, this is summarized in~\autoref{table:1}. Note also that we transformed the annotations of expressions and action units to binary values, where 1 indicates the presence of the expression or action unit, and 0 indicates the absence of the expression or action unit.

\begin{table}[h!]
  \centering
  \caption{Annotation ranges and invalid values.}
  \label{table:1}
  \begin{tabular}{@{}lcc@{}}
    \toprule
    \textbf{Annotation} & \textbf{Range} & \textbf{Invalid} \\ \midrule
    Valence/Arousal         & [0, 1]         & -5                      \\
    Expressions             & \{0, 1\}       & -1                      \\
    Action Units (AUs)      & \{0, 1\}       & -1                      \\ \bottomrule
  \end{tabular}
\end{table}

We applied this rigorous filtering to the train-validation splits, frames containing any annotation values outside the specified acceptable range were excluded from use, even if all other values for that frame were within the normal range. This resulted in a refined dataset of 52,154 frames for training and 15,440 frames for validation, a summary of the dataset after filtering is shown in~\autoref{table:2}.

\begin{table}[h!]
  \centering
  \caption{Dataset summary after filtering}
  \label{table:2}
  \begin{tabular}{@{}lcc@{}}
    \toprule
    \textbf{Partition} & \textbf{Frames} & \textbf{Curated frames} \\ \midrule
    Training                   & 142,382                             & 52,154                                      \\
    Validation                 & 26,876                              & 15,440                                      \\
    Test                       & 52,670                              & -                                           \\ \bottomrule
  \end{tabular}
\end{table}

\subsection{Network architecture}

For the 7th ABAW challenge, we adapted the Dual-Direction Attention Mixed Feature Network (DDAMFN)~\cite{zhang2023ddamfn} to the multitask problem with three different fully-connected layers at the end of the network. These layers consist of a valence-arousal prediction layer with 2 output units, an emotion recognition layer with 8 output units, and an action unit layer with 12 output units.

\begin{figure}[h!]
  \centering
  \includegraphics[width=\textwidth]{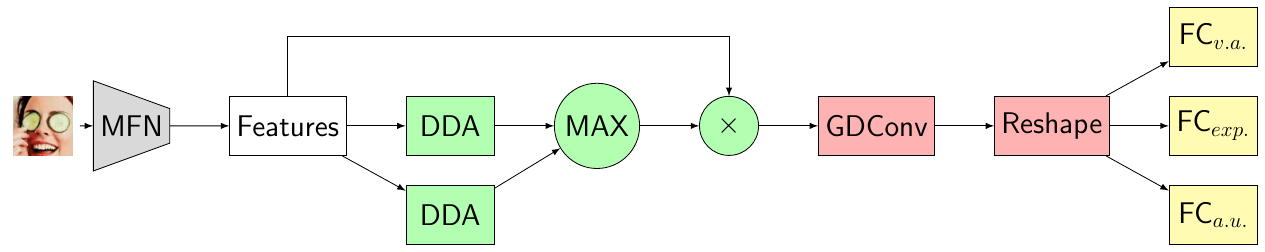}
  \caption{Our DDAMFN~\cite{zhang2023ddamfn} architecture for the 7th ABAW challenge: MobileFaceNet (MFN) for feature extraction (grey), Dual-Direction Attention (DDA) module (green), Global Depthwise Convolution (GDConv) layer (red), and three fully-connected layers for valence-arousal prediction, emotion recognition, and action unit detection (yellow).}
  \label{fig:cnn}
\end{figure}

\autoref{fig:cnn} shows a diagram of the network, which features a base MobileFaceNet (MFN)~\cite{chen2018mobilefacenets} architecture for feature extraction, followed by a Dual-Direction Attention (DDA) module --with two attention heads-- and a Global Depthwise Convolution (GDConv) layer. The output of the GDConv layer is reshaped and fed into the three fully-connected layers for the different tasks.

\subsection{Training}

Initially, the DDAMFN pretrained model was obtained from stock versions in the source code repository of the original work, which was trained on AffectNet-8~\cite{zhang2023ddamfn}. In both our expriments, we preserved the feature extraction, attention mechanisms and the GDConv layer pretained weights, but initialized the fully-connected layers with random weights.

For the first experiment, a multitask training was performed in two different stages. First, we focused on training our custom multitask classifier, so, for training, the entire network was frozen except for the custom classifier. The classifier was trained in isolation to learn the distinct characteristics of each task without interference from the feature extraction layers. Subsequently, the entire model was unfrozen, and fine-tuned on all layers. This comprehensive training stage ensured that the feature extraction, attention mechanisms, and classifier were all optimized to work cohesively for the multitask problem.

Loss functions were calculated following these criteria:

\begin{itemize}
  \item For valence-arousal prediction, the loss function was calculated using the Concordance Correlation Coefficient (CCC). CCC is a measure that evaluates the agreement between two time series by assessing both their precision (how well the observations agree) and accuracy (how well the observations match the true values).
  \item For emotion recognition, was cross-entropy, which is commonly employed in classification tasks to measure the difference between the predicted probability distribution and the true distribution.
  \item For action unit (AU) detection, the binary cross-entropy loss was used, which is suitable for binary classification tasks, measuring the difference between the predicted probability and the actual binary outcome for each action unit.
\end{itemize}

When the whole model was fine-tuned, an extra loss for the attention head was added. This loss works by calculating the mean squared error (MSE) between each pair of attention heads, and then normalizing the total loss by the number of pairs plus a small epsilon to avoid division by zero. This encourages the attention heads to produce consistent outputs \cite{zhang2023ddamfn}.

Furthermore, for the Action Unit task, a global threshold of 0.5 was initially tested across all AUs, followed by individual optimization for each AU~\cite{savchenko2022framelevelpredictionfacialexpressions}.

As for the second experiment, we repeated a similar procedure. However, this time we trained each task individually. This approach allowed us to compare the results for each specific task, providing more detailed insights and enabling a clearer understanding of the performance variations across different tasks.

\section{Results}

The metrics evaluated include the Concordance Correlation Coefficient (CCC) for valence, arousal, and their combination (valence-arousal), F1 score for emotion classification, F1 score for action unit (AU) detection, and the combined performance score (P Score), as described in the challenge~\cite{kollias20247th}. And composed following the challenge P score formulation:

\begin{equation}
  P = \frac{\textrm{CCC}_{V} + \textrm{CCC}_{A}}{2} + \frac{\textrm{F1}_{Expr}}{8} + \frac{\textrm{F1}_{AU}}{12}
\end{equation}

The performance metrics for the multitask challenge are summarized in~\autoref{tab:headings-1}. The fine-tuned DDAMFN model achieved a CCC of 0.549 for valence, 0.524 for arousal, and 0.536 for valence-arousal. The F1 scores for emotion classification and AU detection were 0.277 and 0.470, respectively, resulting in a P score of 1.287. Incorporating threshold adjustments slightly improved the F1 score for AU detection to 0.510 and the P score to 1.327.

Training a custom classifier showed a CCC of 0.548 for valence, 0.518 for arousal, and 0.533 for valence-arousal. The F1 scores for emotion classification and AU detection were 0.262 and 0.473, respectively, with a P score of 1.283. Threshold adjustments for the custom classifier improved the F1 score for AU detection to 0.500 and the P score to 1.313.

\begin{table}[ht]
  \caption{Performance metrics for the multitask challenge. Table shows $\textrm{F1}_{Expr}$ and $\textrm{F1}_{AU}$ already normalized by the number of classes for each classification task. $\textrm{CCC}_{AV}$ is the combined CCC for valence and arousal.}
  \label{tab:headings-1}
  \centering
  \begin{tabular}{@{}lcccccc@{}}
    \toprule
    Trainable layers &  $\textrm{CCC}_{V}$ & $\textrm{CCC}_{A}$ &  $\textrm{CCC}_{VA}$ & $\textrm{F1}_{Expr}$ & $\textrm{F1}_{AU}$ & P \\
    \midrule
    DDAMFN & \textbf{0.549} & \textbf{0.524} & \textbf{0.536} & \textbf{0.277} & 0.470 & 1.287 \\
    DDAMFN+thresholds & \textbf{0.549} & \textbf{0.524} & \textbf{0.536} & \textbf{0.277} & \textbf{0.510} & \textbf{1.327} \\
    Custom classifier & 0.548 & 0.518 & 0.533 & 0.262 & 0.473 & 1.283 \\
    Custom classifier+thresholds & 0.548 & 0.518 & 0.533 & 0.262 & 0.500 & 1.313 \\
    \bottomrule
  \end{tabular}
\end{table}

The performance metrics for individual tasks are detailed in~\autoref{tab:headings-2}. For individual tasks, fine-tuning the entire DDAMFN model demonstrated higher performance with a CCC of 0.604 for valence, 0.550 for arousal, and 0.577 for valence-arousal. The F1 scores for emotion classification and AU detection were 0.287 and 0.490, respectively, with a P Score of 1.354. Threshold adjustments further enhanced the F1 score for AU detection to 0.529 and the P Score to 1.393.

Training a custom classifier presented a CCC of 0.530 for valence, 0.537 for arousal, and 0.533 for valence-arousal. The F1 scores for emotion classification and AU detection were 0.243 and 0.487, respectively, resulting in a P Score of 1.263. Applying threshold adjustments increased the F1 score for AU detection to 0.524 and the P Score to 1.300.

\begin{table}[ht]
  \caption{Performance metrics for individual tasks. Table shows $\textrm{F1}_{Expr}$ and $\textrm{F1}_{AU}$ already normalized by the number of classes for each classification task. $\textrm{CCC}_{AV}$ is the combined CCC for valence and arousal.
  }
  \label{tab:headings-2}
  \centering
  \begin{tabular}{@{}lcccccc@{}}
    \toprule
    Trainable layers &  $\textrm{CCC}_{V}$ & $\textrm{CCC}_{A}$ &  $\textrm{CCC}_{VA}$ & $\textrm{F1}_{Expr}$ & $\textrm{F1}_{AU}$ & P \\
    \midrule
    DDAMFN & \textbf{0.604} & \textbf{0.550} & \textbf{0.577} & \textbf{0.287} & 0.490 & 1.354 \\
    DDAMFN+thresholds & \textbf{0.604} & \textbf{0.550} & \textbf{0.577} & \textbf{0.287} & \textbf{0.529} & \textbf{1.393} \\
    Custom classifier & 0.530 & 0.537 & 0.533 & 0.243 & 0.487 & 1.263 \\
    Custom classifier+thresholds & 0.530 & 0.537 & 0.533 & 0.243 & 0.524 & 1.300 \\
    \bottomrule
  \end{tabular}
\end{table}

Overall, both approaches: fine-tuning the entire DDAMFN model and training a custom classifier, performed well across the evaluated metrics, with fine-tuning the DDAMFN model showing slightly better performance in individual task metrics.

\section{Test set evaluation}
We conducted an extensive evaluation by submitting three distinct sets of prediction models. The first set of predictions was generated using the model that had been fine-tuned across the entire architecture (DDAMFN+thresholds).

The second set of predictions was done leveraging DDAMFN as a feature extractor. In this approach, the sophisticated feature extraction capabilities of DDAMFN were utilized to capture complex patterns and representations within the data. A custom classifier was then trained on these extracted features, allowing for a focused and potentially more accurate classification performance(Custom classifier+thresholds).

The third set of predictions was derived from the ensemble approach, where the results from the models trained individually for each task were combined.

\begin{table}[ht]
  \caption{Performance metrics for the three submissions. Table shows $\textrm{F1}_{Expr}$ and $\textrm{F1}_{AU}$ already normalized by the number of classes for each classification task. $\textrm{CCC}_{AV}$ is the combined CCC for valence and arousal.
  }
  \label{tab:headings-3}
  \centering
  \begin{tabular}{@{}lcccc@{}}
    \toprule
    Models for each submission &  $\textrm{CCC}_{VA}$ & $\textrm{F1}_{Expr}$ & $\textrm{F1}_{AU}$ & P \\
    \midrule
    DDAMFN+thresholds & 0.262 & \textbf{0.282} & 0.478 & 1.022 \\
    Custom classifier+thresholds & \textbf{0.371} & 0.277 & 0.466 & \textbf{1.114} \\
    Ensambled individual model results & 0.292 & 0.272 & \textbf{0.499} & 1.064 \\
    \bottomrule
  \end{tabular}
\end{table}

Results are detailed in ~\autoref{tab:headings-3}. DDAMFN+thresholds: This model achieved the highest F1 score for expressions ($\textrm{F1}{Expr}$ = 0.282), indicating its effectiveness in classifying expressions. However, it had the lowest combined CCC for valence and arousal ($\textrm{CCC}{VA}$ = 0.262), suggesting it is less reliable in predicting emotional valence and arousal. The overall performance metric P was 1.022, positioning it as the least effective in terms of the combined performance metrics.

Custom classifier+thresholds: This model excelled with the highest combined CCC for valence and arousal ($\textrm{CCC}{VA}$ = 0.371), showing its strength in predicting emotional states accurately. It also had the highest overall performance metric P (1.114), reflecting its robust performance across the evaluated metrics. Despite this, its F1 scores for expressions ($\textrm{F1}{Expr}$ = 0.277) and action units ($\textrm{F1}_{AU}$ = 0.466) were slightly lower compared to the other models, indicating room for improvement in those specific areas.

Ensembled individual model results: This model achieved the highest F1 score for action units ($\textrm{F1}{AU}$ = 0.499), making it the best performer in classifying action units. Its combined CCC for valence and arousal was moderate ($\textrm{CCC}{VA}$ = 0.292), and its F1 score for expressions was the lowest among the models ($\textrm{F1}_{Expr}$ = 0.272). The performance metric P was 1.064, indicating a balanced performance but not leading in any single category except for action units.

The model that achieved the highest overall performance was ranked 5th in the 7th Affective Behavior Analysis in the Wild (ABAW) Challenge. This ranking reflects its competitive capability and robustness among numerous submissions from diverse research groups, highlighting its efficacy in real-world affective behavior analysis tasks.

\begin{table}[h!]
  \caption{Multi-Task Learning Challenge Results (7th ABAW)
  }
  \label{tab:headings-4}
  \centering
  \begin{tabular}{@{}lccccc@{}}
    \toprule
    Teams &  $\textrm{CCC}_{VA}$ & $\textrm{F1}_{Expr}$ & $\textrm{F1}_{AU}$ & P \\
    \midrule
    Netease Fuxi AI Lab~\cite{liu2024affectivebehaviouranalysisprogressive} & \textbf{0.5420} & \textbf{0.4286} & \textbf{0.5580} & \textbf{1.5286} \\
    HSEmotion~\cite{savchenko2024hsemotionteam7thabaw} & {0.4074} & 0.3279 & 0.5119 & {1.2472} \\
    HFUT-MAC1~\cite{shen2024facialaffectrecognitionbased} & 0.3783 & 0.2997 & {0.4997} & 1.1777 \\
    SCU-ACers~\cite{li2024affectivebehavioranalysisusing} & 0.3743 & 0.3018 & {0.4879} & 1.1640 \\
    \textit{Ours} & \textit{0.3710} & \textit{0.2772} & \textit{0.4643} & \textit{1.1145} \\
    SML & 0.292 & 0.1938 & {0.4046} & 0.8692 \\
     \midrule
    baseline & 0.1193 & 0.1007 & {0.1200} & 0.3400 \\
    \bottomrule
  \end{tabular}
\end{table}

The key differences compared to other teams are multifaceted. Primarily, other teams have employed more sophisticated and heavier feature extractors, such as Masked Autoencoders (MAE)~\cite{liu2024affectivebehaviouranalysisprogressive}, or have integrated multiple feature extractors to enhance the overall quality of the extracted features~\cite{shen2024facialaffectrecognitionbased}. This approach has allowed them to achieve superior results.

In contrast, our methodology included a more aggressive data filtering process, which consequently led to the utilization of a smaller dataset for training purposes compared to our competitors. This reduction in data volume may have impacted our model's ability to generalize effectively.

Moreover, a significant limitation of our training approach has been the treatment of each frame in isolation, without taking into account any temporal relationships or dependencies within the sequence of frames~\cite{savchenko2024hsemotionteam7thabaw}. This lack of temporal context has posed challenges for the training process, potentially hindering the model's performance in tasks where temporal dynamics are crucial.

Another notable distinction lies in our reliance on a single, streamlined model architecture. Unlike some other teams, which have explored the benefits of ensemble methods or hybrid models incorporating both classical and deep learning techniques, our approach has remained relatively straightforward~\cite{li2024affectivebehavioranalysisusing}.

Overall, while our methods have certain strengths, these differences in feature extraction, data volume, and temporal context consideration have been critical factors influencing our comparative performance.

\section{Conclusion}

The results using DDAMFN as a feature extractor with the pretrained weigths were practically the same as those obtained by fine-tuning the entire model. This similarity in performance can be attributed to the architecture of DDAMFN, which has been trained to generalize exceptionally well for the multitask challenge problem. This inherent capability allows it to extract meaningful features that are sufficient for achieving high performance without the need for extensive fine-tuning.

In contrast, the single-task experiment revealed a different scenario. Fine-tuning the entire model for the single-task yielded better results than merely training the classifier, and even outperformed the multitask results. This finding underscored the importance of task-specific optimization. When the model is fine-tuned for a specific task, it can leverage the nuances and particularities of that task, leading to superior performance.

This suggests that with meticulous optimization of the loss functions and careful consideration of data imbalance, the DDAMFN architecture could potentially achieve even better results in the multitask challenge. Proper handling of these aspects could unlock the full potential of the model, leading to significant performance gains in multitask settings.

Moreover, it is crucial to highlight the importance of effective threshold optimization in the Action Units task. By fine-tuning the thresholds, the results improved significantly, with an increase of 0.5 in performance metrics. This improvement demonstrates that beyond model architecture and training strategies, the post-processing steps such as threshold optimization play a vital role in achieving optimal results. Effective threshold optimization ensures that the model's predictions are more accurate and reliable, contributing to overall performance enhancement.

Interestingly, although the least performant model on the evaluation set was the Custom classifier + thresholds, it achieved the best score on the test set. This suggests that the Custom classifier + thresholds might have better generalization capabilities, highlighting the importance of thorough evaluation on diverse test sets.

In summary, while DDAMFN as a feature extractor performs on par with full model fine-tuning in a multitask setting, the single-task fine-tuning shows that there is room for improvement with targeted optimization strategies in multitask scenarios. This includes better loss function optimization, addressing data imbalance, and refining threshold settings, all of which are crucial for maximizing the performance of the model in the multitask challenge. 

\section*{Acknowledgments}

 This work is part of the project PLEC2021-007868, funded by MCIN and by the EU, and the project PID2022-138721NB-I00, funded by MCIN.


%
%
\bibliographystyle{splncs04}
\bibliography{main}
\end{document}